\documentclass{ws-procs9x6}
\usepackage{textcomp}
\usepackage{enumitem}

\begin{document}

\title{ON THE REDUCTION OF BIASES IN BIG DATA SETS FOR THE DETECTION OF IRREGULAR POWER USAGE}

\author{PATRICK GLAUNER and RADU STATE}

\address{Interdisciplinary Centre for Security, Reliability and Trust, University of Luxembourg\\ 
1855 Luxembourg, Luxembourg \\
Email: \{patrick.glauner, radu.state\}@uni.lu
}

\author{PETKO VALTCHEV}

\address{Department of Computer Science, University of Quebec in Montreal\\ 
H3C 3P8 Montreal, Canada \\
Email: valtchev.petko@uqam.ca
}

\author{DIOGO DUARTE}

\address{CHOICE Technologies Holding S\`arl\\ 
2453 Luxembourg, Luxembourg \\
Email: diogo.duarte@choiceholding.com
}

\begin{abstract}
In machine learning, a bias occurs whenever training sets are not representative for the test data, which results in unreliable models. The most common biases in data are arguably class imbalance and covariate shift. In this work, we aim to shed light on this topic in order to increase the overall attention to this issue in the field of machine learning.
We propose a scalable novel framework for reducing multiple biases in high-dimensional data sets in order to train more reliable predictors. We apply our methodology to the detection of irregular power usage from real, noisy industrial data. In emerging markets, irregular power usage, and electricity theft in particular, may range up to 40\% of the total electricity distributed. Biased data sets are of particular issue in this domain. We show that reducing these biases increases the accuracy of the trained predictors. Our models have the potential to generate significant economic value in a real world application, as they are being deployed in a commercial software for the detection of irregular power usage.
\end{abstract}


\bodymatter

\section{Introduction}
\begin{figure}[h!]
    \centering
    \includegraphics[width=0.8\textwidth]{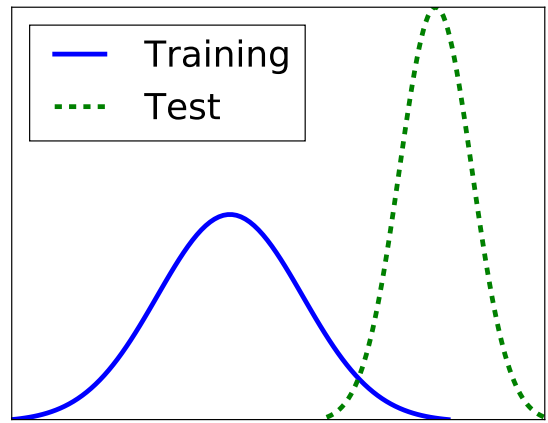}
    \caption{Example of covariate shift: training and test data having different distributions.}
    \label{fig:biasexample}
\end{figure}

The contemporary Big Data paradigm can be summarized as follows: ``It's not who has the best algorithm that wins. It's who has the most data." \cite{banko2001scaling}
However, in many cases, increasing the amounts of data is not a panacea since it can be biased: One frequently appearing bias results in training data and test data having different distributions, as depicted in Fig.~\ref{fig:biasexample}. Learning from such training data leads to unreliable predictors that are not able to generalize to the test data. In the literature, this sort of bias is called covariate shift, sampling bias or sample selection bias. Covariate shift has been recognized as an issue in statistics since the mid-20th century \cite{2017glaunerisbigdata}.
In contrast, it has received only a limited attention in machine learning, mainly within the computational learning theory subfield, yet the situation is currently evolving \cite{shimodaira2000improving, cortes2014domain}. 

Non-technical losses (NTL) appear in power grids during distribution and include, but are not limited to, the following causes: meter tampering in order to record lower consumptions, bypassing meters by rigging lines from the power source, arranged false meter readings by bribing meter readers or faulty or broken meters. NTL are more common in emerging countries, where electricity theft is the main contributor. NTL are reported to range up to 40\% of the total electricity distributed in countries such as Brazil, India, Malaysia or Pakistan \cite{glauner2017challenge, viegas2017solutions}. NTL are the source of major concerns for the electricity providers including financial losses and a decrease of stability and reliability in power grids. It is therefore crucial to detect customers that cause NTL. Recent research on NTL detection mainly uses machine learning models that learn anomalous behavior from customer data and known irregular behavior that was reported through on-site inspection results.
In order to detect NTL more accurately, one may assume that having simply more customer and inspection data would help. We have previously shown that in many cases, the set of inspected customers is biased \cite{2017glaunerisbigdata}. A reason for that is that past inspections have been largely focused on certain criteria and were not sufficiently spread across the population.

This paper builds on top of our previous contributions and aims at bias reduction in data, and further at more generalizable NTL predictors. Its main contributions are:
\begin{itemize}
\item We present a framework for reducing biases in data, such as class imbalance and covariate shift, in particular for spatial data.
\item We propose a scalable novel methodology for reducing multiple biases in high-dimensional data sets at the same time.
\item We report on how our method performs on the detection of NTL. Our method leads to a better detection of anomalous customers, subsequently reduces losses of electricity providers and thus increases stability and reliability of power distribution infrastructure.
\end{itemize}

\section{Background and Related Work}
In supervised learning, training examples $(x^{(i)}, y^{(i)})$ are drawn from a training distribution $P_{train}(X, Y)$, where $X$ denotes the data and $Y$ the label, respectively. The training set is biased if $P_{train}(X, Y)\ne P_{test}(X, Y)$. In order to reduce the bias, it has been shown that example $(x^{(i)}, y^{(i)})$ can be weighted during training as follows\cite{jiang2008literature}:

\begin{align*}
w_i = \frac{P_{test}(x^{(i)}, y^{(i)})}{P_{train}(x^{(i)}, y^{(i)})}.
\end{align*}
However, computing $P_{train}(x^{(i)}, y^{(i)})$ is impractical because of the limited amount of data in the training domain. It is for that reason that in the literature, predominantly two different types of biases are discussed: class imbalance and covariate shift. Class imbalance refers to the case where classes are unequally represented in the data. Therefore, we assume $P_{train}(X\lvert Y) = P_{test}(X\lvert Y)$, but $P_{train}(Y)\ne P_{test}(Y)$. \cite{japkowicz2002class} In contrast, for covariate shift, we assume $P_{train}(Y\lvert X) = P_{test}(Y\lvert X)$, but $P_{train}(X)\ne P_{test}(X)$. \cite{zadrozny2004learning}. Instance weighting using density estimation has been proposed for correcting covariate shift. \cite{shimodaira2000improving}
Furthermore, the Heckman method has been proposed to correct covariate shift. \cite{10.2307/1912352} However, the Heckman method only applies to logistic regression models. 
Other biases are reported in the literature, for example for change of functional relations, i.e. when $P_{train}(Y\lvert X) \ne P_{test}(Y\lvert X)$, or biases created by transforming the feature space. \cite{jiang2008literature}

\section{Reduction of Biases}
\label{chapter:methodology}
We propose the following methodology: Given the assumptions made for class imbalance, we compute the corresponding weight for example $i$ having a label of class $k$ as follows:
\begin{align*}
w_{i,k} = \frac{P_{test}(x^{(i)}, y_k^{(i)})}{P_{train}(x^{(i)}, y_k^{(i)})} = \frac{P_{test}(x^{(i)}\lvert y_k^{(i)})P_{test}(y_k^{(i)})}{P_{train}(x^{(i)}\lvert y_k^{(i)})P_{train}(y_k^{(i)})} =\frac{P_{test}(y_k^{(i)})}{P_{train}(y_k^{(i)})}.
\end{align*}
We use the empirical counts of classes for computing $P_{<dist>}(y_k)$.
Given the assumptions made for covariate shift, we compute the corresponding weight for the bias in feature $k$ of example $i$ as follows:
\begin{align*}
w_{i,k} = \frac{P_{test}(x_k^{(i)}, y^{(i)})}{P_{train}(x_k^{(i)}, y^{(i)})} = \frac{P_{test}(y^{(i)}\lvert x_k^{(i)})P_{test}(x_k^{(i)})}{P_{train}(y^{(i)}\lvert x_k^{(i)})P_{train}(x_k^{(i)})} = \frac{P_{test}(x_k^{(i)})}{P_{train}(x_k^{(i)})}.
\end{align*}
We use density estimation for computing $P_{<dist>}(x_k^{(i)})$\cite{scikit-learn}.

There may be a variety of biases in a learning problem that are far more than just class imbalance and covariate shift on a single dimension. We have shown previously that there may be multiple types of covariate shift, for example spatial covariate shifts on different hierarchical levels. There may be also covariate shifts for other master data, such as for the customer class or for the contract status \cite{2017glaunerisbigdata}.
We now aim to correct $n$ different biases at a same time, e.g. for class imbalance as well as different types of covariate shift. As $x^{(i)}$ has potentially many dimensions with a considerable covariate shift, computing the joint $P_{<dist>}(x^{(i)})$ becomes impractical for an increasing number of dimensions. We propose a uniformed and scalable solution to combine weights for correcting the $n$ different biases, comprising for example of class imbalance and different types of covariate shift. The corresponding weights per bias of an example are $w_{i,1}, w_{i,2}, ..., w_{i,n}$. The example weight $w_i$ is the harmonic mean of the weights of the biases considered is computed as follows:
\begin{align}
w_i = \frac{n}{\frac1{w_{i,1}} + \frac1{w_{i,2}} + \cdots + \frac1{w_{i,n}}} = \frac{n}{\sum\limits_{k=1}^n \frac1{w_{i,k}}}.
\label{eq:main}
\end{align}

As the different $w_{i,k}$ are computed from noisy, real-world data, special care needs to be paid to outliers. Outliers can potentially lead to very large values $w_{i,k}$ for the density estimation proposed above. It is for that reason that we choose the harmonic mean, as it allows to penalizes extreme values and give preference to smaller values.

%Furthermore, in some use cases, certain biases may be more important to reduce than others. Therefore, we propose that bias $k$ can be associated a factor $f_k$ by computing the weighted (in that context: factored) harmonic mean:
%\begin{align}
%w_i = \frac{\sum\limits_{k=1}^n f_k}{\sum\limits_{k=1}^n \frac{f_k}{w_{i,k}}}.
%\end{align}

\section{Evaluation}
\label{chapter:eval}
The data used in this paper comes from an electricity provider in Brazil, from which we retain $M=150,700$ customers. For these customers, we have a complete time series of 24 monthly meter readings before the most recent inspection. From each time series, we compute 304 features comprising generic time series features, daily average features and difference features, as detailed in Table~\ref{table:features}. The computation of these features is explained in detail in our previous work\cite{2017glauneridentifying}.

\begin{table}[h!]
\tbl{Number of features before and after selection.}
{\begin{tabular}{l c c}\toprule
Name & \#Features & \#Retained features \\
\colrule
Daily average & 23 & 18 \\
Fixed interval & 36 & 34 \\
Generic time series & 222 & 162 \\
Intra year difference & 12 & 12 \\
Intra year seasonal difference & 11 & 11 \\
\colrule
Total & 304 & 237 \\
\botrule
\end{tabular}}
\label{table:features}
\end{table}

Next, we employ hypothesis tests to the features in order to retain the ones that are statistically relevant. These tests are based on the assumption that a feature $x_k$ is meaningful for the prediction of the binary label vector $y$ if $x_k$ and $y$ are not statistically independent \cite{radivojac2004feature}. For binary features, we use Fisher's exact test \cite{fisher1922interpretation}. In contrast, for continuous features, we use the Kolmogorov-Smirnov test \cite{massey1951kolmogorov}. We retain 237 of the 304 features.

We previously found a random forest (RF) classifier to perform the best on this data compared to decision tree, gradient boosted tree and support vector machine classifiers\cite{2017glauneridentifying}. It is for this reason that in the following experiments, we only train RF classifiers. When training a RF, we perform model selection by doing randomized grid search, for which the parameters are detailed in Table~\ref{table:rf}. We use 100 sampled models and perform 10-fold cross-validation for each model.

\begin{table}[h!]
\tbl{Model parameters for random forest.}
{\begin{tabular}{l c}\toprule
Parameter & Values \\
\colrule
Max. number of leaves & $[2, 1000)$ \\
Max. number of levels & $[1, 50)$ \\
Measure of the purity of a split & $\{$entropy, gini$\}$ \\
Min. number of samples required to be at a leaf & $[1, 1000)$ \\
Min. number of samples required to split a node & $[2, 50)$ \\
Number of estimators & $20$ \\
\botrule
\end{tabular}}
\label{table:rf}
\end{table}

We have previously shown that the location and class of customers have the strongest covariate shift \cite{2017glaunerisbigdata}. When reducing these, we first compute the weights for the class imbalance, the spatial covariate shift and customer class covariate shift, respectively, as defined in Sec.~\ref{chapter:methodology}. For covariate shift, we use randomized grid search for a model selection of the density estimator that is composed of the kernel type and kernel bandwidth. The complete list of parameters and considered values is depicted in Table~\ref{table:config}.

\begin{table}[h!]
\tbl{Density estimation parameters.}
{\begin{tabular}{@{}lc@{}}\toprule
Parameter & Values \\
\colrule
Kernel & $\{$gaussian, tophat, epanechnikov, exponential, linear, cosine$\}$ \\
Bandwidth & $[0.001, 10]$ (log space) \\
\botrule
\end{tabular}}
\label{table:config}
\end{table}

Next, we use Eq.~\ref{eq:main} to combine these weights step by step. For each step, we report the test performance of the NTL classifier in Table~\ref{table:results}. It clearly shows that the larger the number of addressed biases, the higher the reliability of the learned predictor.

\begin{table}[h!]
\tbl{Test performance of random forest.}
{\begin{tabular}{@{}lc@{}}\toprule
Biases reduced & $\overline{AUC}$ \\
\colrule
None & 0.59535 \\
Class imbalance & 0.64445 \\
Class imbalance + spatial covariate shift & 0.71431 \\
Class imbalance + spatial covariate shift  + customer class covariate shift & 0.73980 \\
\botrule
\end{tabular}}
%\begin{tabnote}
\textit{Note}: We use the area under the receiver-operating curve (AUC) metric. It is particularly useful for NTL detection, as it allows to handle imbalanced datasets and puts correct and incorrect inspection results in relation to each other\cite{glauner2017challenge}. $\overline{AUC}$ denotes the mean test AUC of the 10 folds of cross-validation for the best model.\\
%\end{tabnote}
\label{table:results}
\end{table}

% For spatial: Bandwidth = 0.69519, Gaussian

\section{Conclusions and Future Work}
Biases appear in many real-world applications of machine learning and refer to the training data not being representative for the test data. The most common biases are class imbalance and covariate shift. In this work, we proposed a scalable model for reducing multiple biases in high-dimensional data at the same time. We applied our methodology to a real-world, noisy data set on irregular power usage. Our model leads to more reliable predictors, thus allowing to better detect customers that have an irregular power usage.
Next, we aim to evaluate our methodology on other data sets, to derive models that reduce hierarchical spatial biases and to handpick a unbiased test set as ground truth for evaluation.

\section*{Acknowledgement}
The present project is supported by the National Research Fund, Luxembourg under grant agreement number 11508593.

\bibliographystyle{ws-procs9x6}
\bibliography{references}

\end{document}